%% file: collas2025_conference__1_.tex
\documentclass{article} 
\usepackage{collas2025_conference,times}
\usepackage{easyReview}

\input{math_commands.tex}

\usepackage{algorithm}
\usepackage{algorithmic}
\usepackage{booktabs} 
\usepackage{subfigure}
\expandafter\def\csname ver@subfig.sty\endcsname{}

\usepackage{svg}
\usepackage{amsmath}
\usepackage{amssymb}
\usepackage{mathtools}
\usepackage{amsthm}
\usepackage{multirow}
\usepackage{titlesec} 
\usepackage{listings}
\usepackage{xcolor}

\usepackage{xcolor}
\usepackage[normalem]{ulem} 

\newif\ifshowrevisions
\showrevisionstrue

\ifshowrevisions
    \newcommand{\added}[1]{\textcolor{blue}{#1}}
    \newcommand{\deleted}[1]{\textcolor{red}{\sout{#1}}}
\else
    \newcommand{\added}[1]{#1}
    \newcommand{\deleted}[1]{}
\fi

\definecolor{white}{rgb}{1,1,1}
\definecolor{bg}{rgb}{0.95,0.95,0.95}
\usepackage{hyperref}
\hypersetup{
    colorlinks=true,
    linkcolor=red,
    filecolor=magenta,
    urlcolor=blue,
    citecolor=purple,
    pdftitle={Overleaf Example},
    pdfpagemode=FullScreen,
    }

\title{Mitigating the Stability-Plasticity Dilemma in Adaptive Train Scheduling with Curriculum-Driven Continual DQN Expansion}

\collasfinalcopy

\showrevisionstrue

\begin{document}

\author{
    Achref Jaziri\textsuperscript{\rm 1}\thanks{Equal contribution.} , 
     Étienne Künzel\textsuperscript{\rm 1}\footnotemark[1] , 
    Visvanathan Ramesh\textsuperscript{\rm 1,2}  \\
    \textsuperscript{\rm 1} Department of Computer Science and Mathematics, Goethe University, \\
    \textsuperscript{\rm 2} HessianAI, Frankfurt am Main, Germany \\
    \{jaziri, vramesh\}@em.uni-frankfurt.de
}

\maketitle

\begin{abstract}
A continual learning agent builds on previous experiences to develop increasingly complex behaviors by adapting to non-stationary and dynamic environments while preserving previously acquired knowledge. However, scaling these systems presents significant challenges, particularly in balancing the preservation of previous policies with the adaptation of new ones to current environments. This balance, known as the stability-plasticity dilemma, is especially pronounced in complex multi-agent domains such as the train scheduling problem, where environmental and agent behaviors are constantly changing, and the search space is vast. In this work, we propose addressing these challenges in the train scheduling problem using curriculum learning. We design a curriculum with adjacent skills that build on each other to improve generalization performance. Introducing a curriculum with distinct tasks introduces non-stationarity, which we address by proposing a new algorithm: Continual Deep Q-Network (DQN) Expansion (CDE). Our approach dynamically generates and adjusts Q-function subspaces to handle environmental changes and task requirements. CDE mitigates catastrophic forgetting through EWC while ensuring high plasticity using adaptive rational  activation functions. Experimental results demonstrate significant improvements in learning efficiency and adaptability compared to RL baselines and other adapted methods for continual learning, highlighting the potential of our method in managing the stability-plasticity dilemma in the adaptive train scheduling setting.
\end{abstract}

\section{Introduction}
Continual learning agents, capable of acquiring new knowledge and skills over time, are essential for environments that constantly change \cite{abel2024definition,khetarpal2022towards}. One central challenge for continual learning systems is balancing the retention of previously learned knowledge with the acquisition of new knowledge—a concept known as the stability-plasticity dilemma \cite{schwarz2018progress}. This dilemma is relevant in many application domains such as the train scheduling problem (TSP), where environmental conditions (i.e. routes, schedules) and agent behaviors (i.e. train characteristics) are constantly changing, and the search space is vast. TSP is inherently complex. Even in simplified scenarios, agents often require considerable time to converge to optimal solutions \cite{toth2002vehicle}.

\looseness=-1
One approach to address the complexity of TSP is curriculum learning, which introduces a sequence of environments that gradually increase in difficulty, allowing the agent to build upon previously acquired skills \cite{narvekar2020curriculum}. This approach mirrors how humans learn, starting with simpler concepts and progressively tackling more complex ones. However, curriculum learning introduces its own set of challenges regarding non-stationary learning. 

\looseness=-1
Various continual reinforcement learning (CRL) approaches have been proposed to mitigate catastrophic forgetting, improve plasticity \cite{klein2024plasticity}, facilitate transfer learning across new tasks, scale effectively to handle multiple tasks and address the stability-plasticity dilemma \cite{riemerlearning, kim2023stability}. Current methods can be categorized into fixed-size approaches, which do not change the size of the network for new tasks (i.e., regularization-based approaches or optimization-based methods), and network expansion approaches. While network expansion approaches are more likely to achieve better performance, they may suffer from scaling issues as the number of tasks increases \cite{wang2023comprehensive}.

To address this, recent studies in the mode connectivity literature have investigated whether multitask and continual solutions are connected by a manifold of low error. Results show that different optima obtained by gradient-based optimization methods are connected by simple paths of non-increasing loss \cite{mirzadeh2020linear}. Inspired by this, various works have demonstrated how to efficiently build simplicial complexes for fast ensembling and adaptively grow the subspace as the system encounters new tasks. This can be leveraged to develop neural systems that scale better while maintaining good performance across multiple tasks \cite{benton2021loss,gaya2022building}.

\looseness=-1

While previous works focused on continual learning with distinct tasks \added{\cite{kim2023achieving,kim2023stability}}, a curriculum does not involve distinct tasks with clear boundaries but rather a series of adjacent skills that build on top of each other, such as pathfinding, dealing with unexpected malfunctions, or avoiding deadlocks in TSP. This introduces a unique type of non-stationarity, where agents are learning interrelated skills. The challenge lies in effectively sequencing these skills to maximize learning efficiency and performance while also addressing the stability-plasticity dilemma for the learning agents.

\looseness=-1

\begin{figure}[t]
	\centering
	\includegraphics[width=0.6\linewidth]{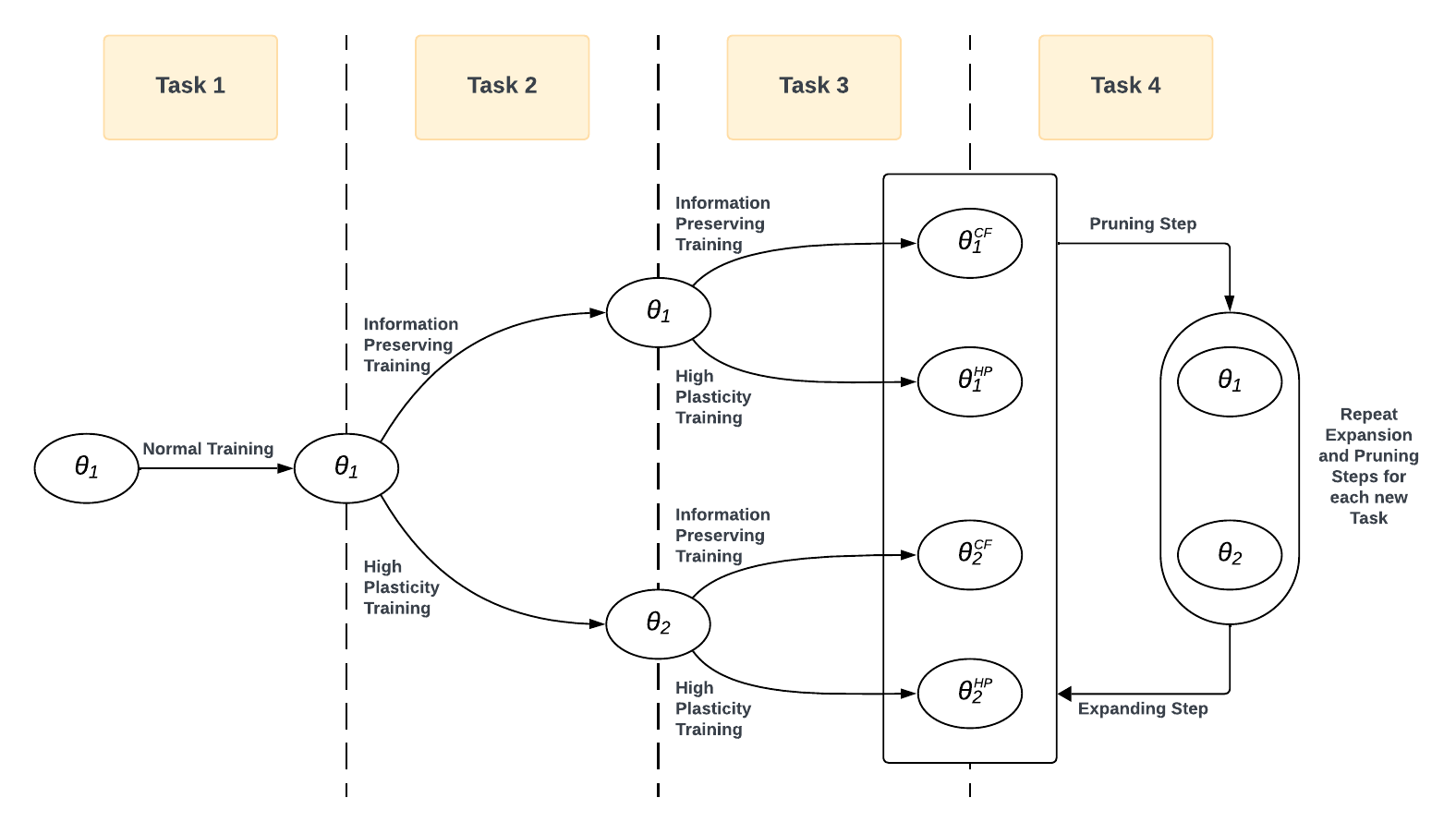} \hspace{0.001cm}
   
	    \caption{Diagram illustrating the Continual DQN Expansion Algorithm. When a new task or a change in the environment occurs, a new subspace with anchor $\theta$ is initialized. The new Q-function is trained with adaptive rational activation functions ($\theta^{HP}_i$) to enhance plasticity and achieve faster convergence on the new task. Simultaneously, the previous subspaces are further adapted on the new task by training Q-functions ($\theta^{CF}_i$) using elastic weight consolidation (EWC) regularization to mitigate catastrophic forgetting on previous tasks. Depending on the performance of different subspaces, the subspace set is either extended or pruned.}

	\label{CDE_Diagram}
\end{figure}

In this work, we propose a well-structured curriculum for the train scheduling problem, decomposing it into various skills like pathfinding, malfunction handling, and deadlock avoidance. Furthermore, we investigate different approaches to leverage this curriculum and introduce the Continual Deep Q-Network Expansion (CDE) algorithm (Figure \ref{CDE_Diagram}). The core idea of our CDE approach is to dynamically expand, adjust, and prune Q-function subspaces based on current task changes in the environment. The decision-making process within CDE involves determining whether a pre-existing Q-function subspace is effective for a new task and thus adapting it, or if a new subspace needs to be created for faster learning. The pre-existing Q-function subspaces are adapted to new tasks with elastic weight consolidation to avoid catastrophic forgetting. while the newly created subspaces are trained with adaptive rational  activation functions for higher plasticity. In the pruning step, we retain only the best Q-networks to maintain sublinear scaling. Experimental results demonstrate significant improvements compared to other approaches, highlighting the potential of CDE to balance stability and plasticity, leading to more efficient and adaptable learning\footnote{A link to the code will be added in the camera ready version.}.

\section{Related Work}
\textbf{Continual Learning:}
Traditional RL methods, which assume static environments, often lead to models prone to overfitting and lacking adaptability to new, varying contexts. Continual RL (CRL)  aims to develop adaptive models that learn from a continuous data stream and evolve across multiple tasks by leveraging accumulated knowledge, thereby overcoming the limitations of static-task assumptions \cite{abel2024definition}. 

\looseness=-1
\textbf{Addressing Catastrophic Forgetting:} Efforts such as those by \citet{rusu2016progressive}, \citet{cheung2019superposition}, and \citet{wortsman2020supermasks} aim to preserve model parameters from previously learned tasks but often do not scale well with an increasing number of tasks mainly due to substantial computational and memory requirements. Other strategies, like maintaining a memory buffer of prior experiences \cite{lopez2017gradient,isele2018selective,caccia2020online}, also face scalability issues as memory costs increase with task complexity and number. Methods enhancing transfer to new tasks, such as fine-tuning, are scalable and adaptable but prone to catastrophic forgetting. Elastic weight consolidation  \cite{kirkpatrick2017overcoming} addresses this by limiting weight modifications, although it may reduce the network plasticity as the number of tasks increases. Knowledge distillation techniques, as discussed in \cite{rusu2015policy,li2017learning,rolnick2019experience}, and methods leveraging similarities across tasks \cite{xu2018reinforced,pasunuru2019continual}, have been utilized to improve CRL performance. Meta-learning and generative models also enhance adaptability and learning efficiency \cite{javed2019meta,beaulieu2020learning,shin2017continual}. However, practical application challenges persist, as indicated by \cite{lee2020neural}, who propose an iteratively expanding network model primarily for supervised learning without distinct task boundaries.

\textbf{Addressing Neural Plasticity:} While addressing catastrophic forgetting is essential for non-stationary agents, extensive neural plasticity is required due to inherent distribution shifts during agent training, such as input and output drifts \cite{mnih2015human}. These shifts are exemplified by various environment dynamics as agents adapt to changing states and rewards. Recent studies have explored approaches to increase plasticity and allow faster adaptation to changing environments, such as concatenated ReLU \cite{abbas2023loss} and the introduction of new trainable weights \cite{nikishin2024deep}. Additionally, techniques like continual backpropagation \cite{dohare2023maintaining} and dynamic hyperparameter and activation function adaptation \cite{loni2023learning,delfosse2021adaptive} highlight the potential of dynamically altering the weight optimization landscape.

\textbf{Curriculum Learning:}
Curriculum learning arranges training data in gradually increasing difficulty to introduce concepts incrementally and improve learning outcomes \cite{bengio2009curriculum}. It has demonstrated enhanced sample efficiency and convergence across various machine learning applications \cite{soviany2022curriculum, wang2021survey}. In reinforcement learning (RL), curriculum learning can boost efficiency and performance \cite{narvekar2020curriculum}. Early RL curricula were manually designed by domain experts, for instance in \citet{Stanley2005}, where video game agents were trained via expert-crafted task sequences.

Recent work has shifted toward automated curriculum generation. For example, \citet{Narvekar2019} introduced a Curriculum Markov Decision Process that selects subsequent tasks based on learning progress. Other approaches cast curriculum sequencing as a combinatorial optimization problem, using metaheuristics like beam search or genetic algorithms to identify optimal task orders \cite{Foglino2019a, Foglino2019b}. \citet{florensa2017reverse} proposed “reverse” training, starting from goal states and progressively moving to initial states farther from the goal, thus automatically adapting the curriculum to the agent’s performance. Nevertheless, learning a separate curriculum policy for each agent and task is computationally expensive \cite{Narvekar2019}, and catastrophic forgetting can occur when transitioning between tasks. Furthermore, when tasks differ in state and action spaces, transferring high-level knowledge (e.g., partial policies or options) becomes critical \cite{Yang1996, Zimmer2018}.

\begin{figure*}[t]
    \centering
    \subfigure[]{\includegraphics[height=1.6cm]{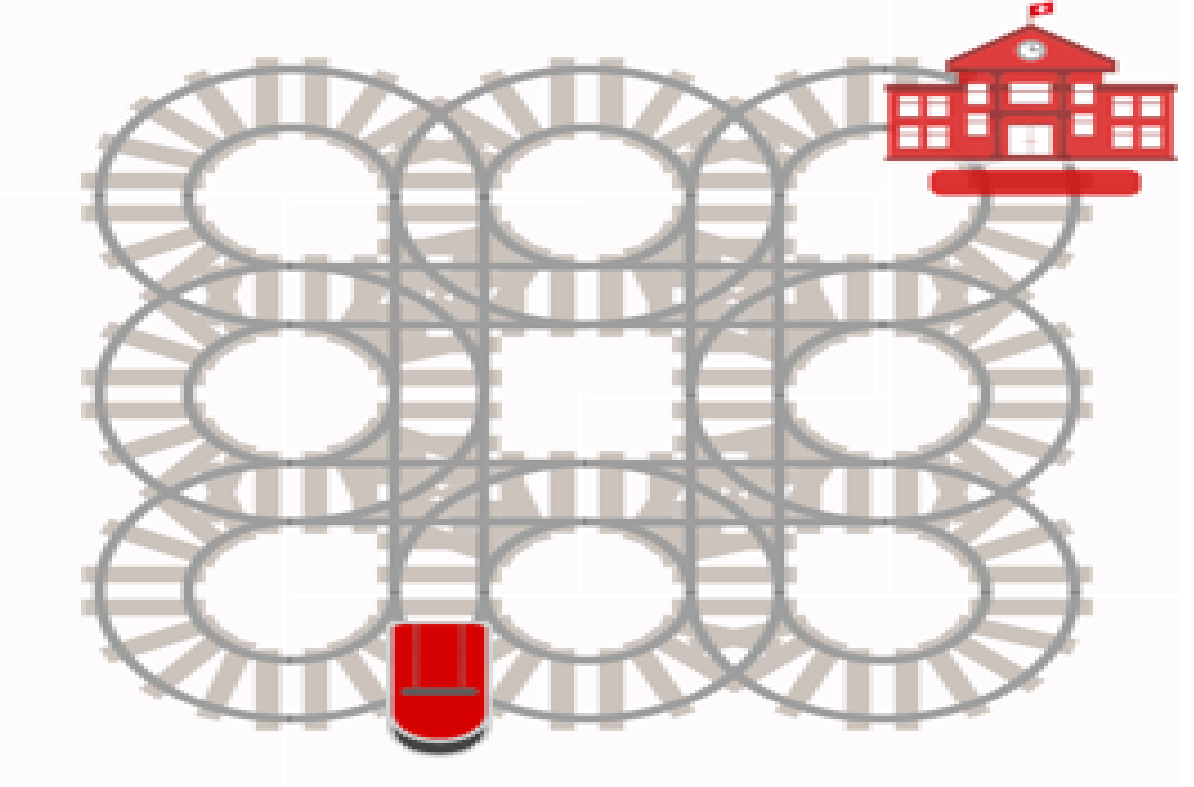}}
    \hspace{0.01cm}
    \subfigure[]{\includegraphics[height=1.6cm]{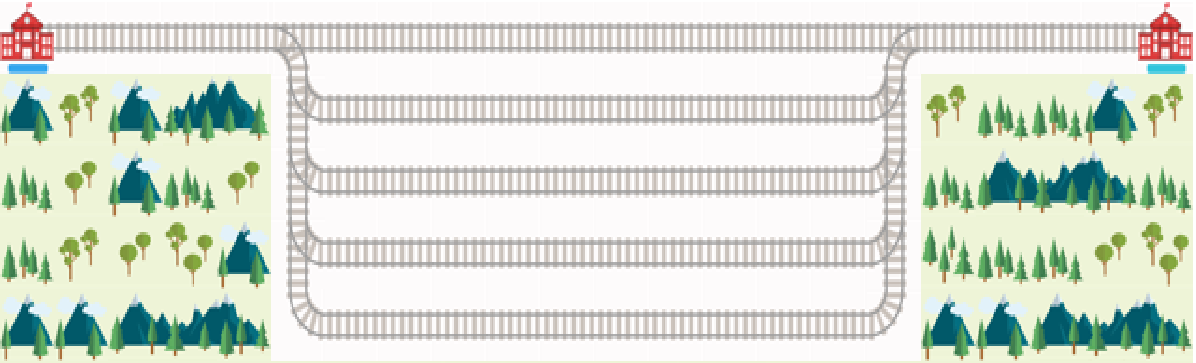}}
    \hspace{0.01cm}
    \subfigure[]{\includegraphics[height=1.6cm]{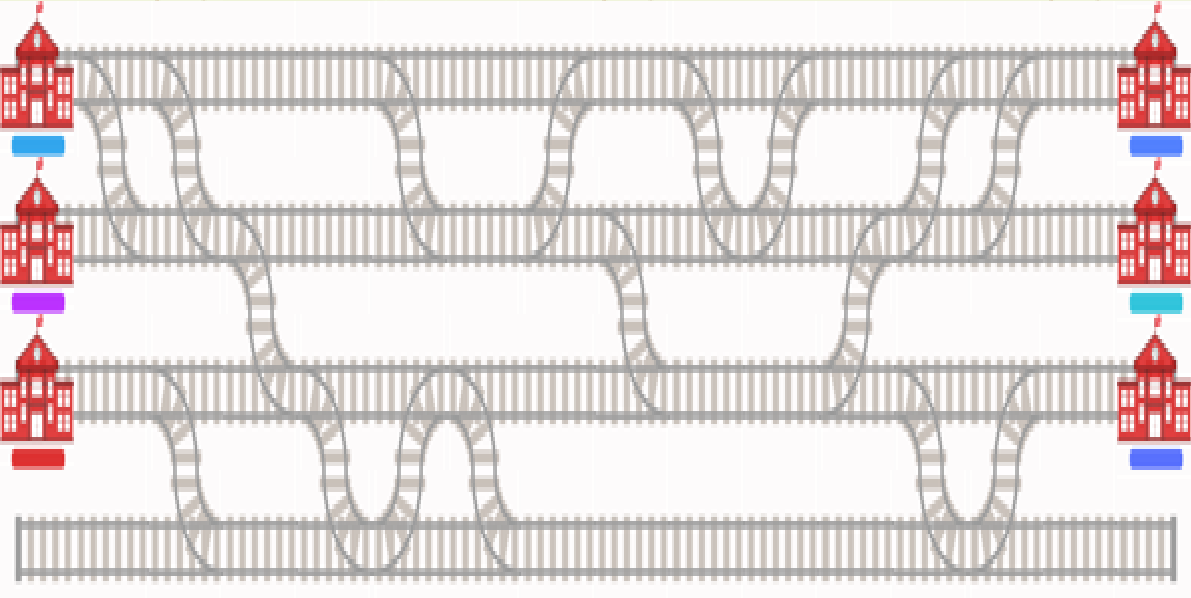}}
    \hspace{0.01cm}
    \subfigure[]{\includegraphics[height=1.6cm]{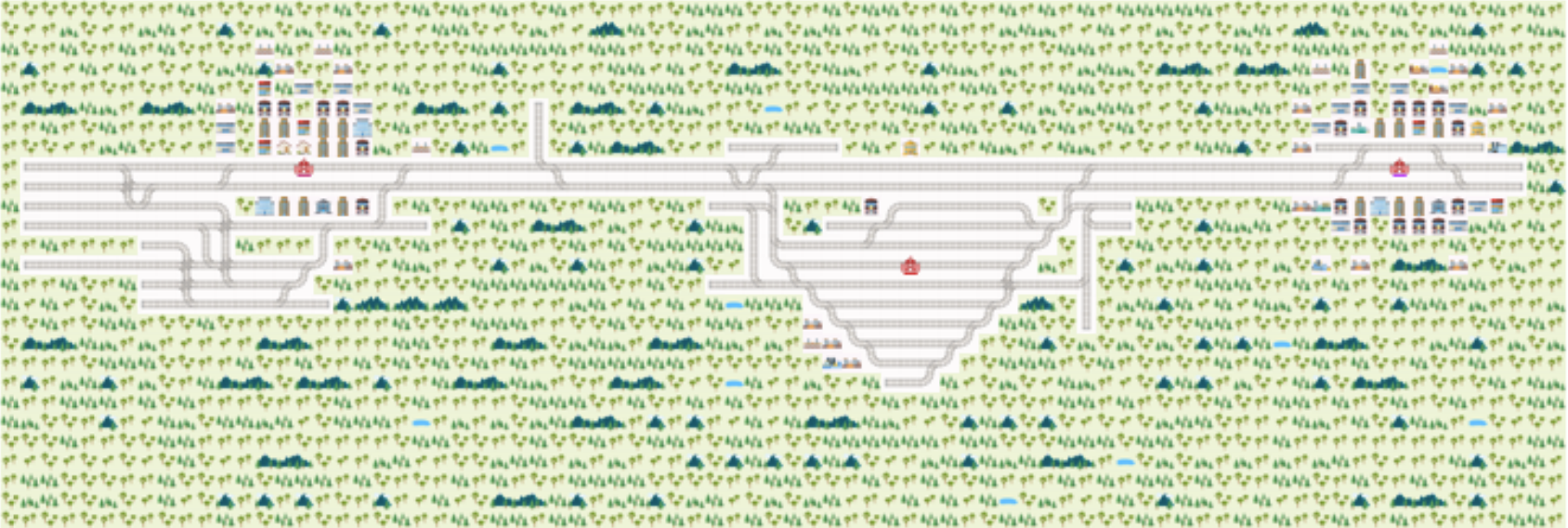}}
    \caption{Illustration of the curriculum designed for TSP through distinct environments. (a) \textbf{Pathfinding Environment}: A grid where a single agent navigates from start to destination, emphasizing efficient route planning. (b) \textbf{Malfunctions and Train Speeds Environment}: A setting where agents deal with random malfunctions and varying train speeds, requiring adaptive responses and strategic planning to minimize delays. (c) \textbf{Deadlocks Environment}: Scenarios focusing on common deadlock situations, gradually increasing in complexity with more agents and fewer switches, to train agents in avoiding gridlock. 
(d) \textbf{Full Task Evaluation Environment}: A fixed environment which emulates a realistic railway network.
The curriculum environments (a-c) are available in multiple sizes to provide diverse training scenarios, as detailed in the appendix. This curriculum structure improves learning efficiency and performance on the full task evaluation environment (d).}
    \label{fig:main}
\end{figure*}
\section{Train Scheduling Problem}
\subsection{Mathematical Formulation}
\looseness=-1

TSP is an NP-Hard optimization problem focused on the optimal assignment of trains to routes and their scheduling across a railway network while adhering to diverse constraints \cite{caprara2002modeling}. It requires dynamic replanning to adapt to unforeseen disruptions. This issue integrates aspects of Multi-Agent Pathfinding and the Vehicle Rescheduling Problem \cite{cai1994fast}.

Consider a network of yards indexed from $0$ to $v$. Each train is assigned a route \( (i,j) \) where $i$ and $j$ are yards in the network. Trains have varying speeds and malfunction rates. In the case of a malfunction, a train is halted for a certain number of time steps.

Given $N_{\text{tr}}$ (number of trains), $V = \{1, \ldots, v\}$   (set of yards), $E_r$ (stop-schedule for train $r$), $v_k$ (speed of train $r$), $m_r$ (malfunction rate of train $r$), $h_r$ (time steps train $r$ is halted in case of a malfunction), $d_{ij}$ (distance between yards $i$ and $j$), $x_{ij}^r$ (binary variable  indicating if train $r$ travels directly from yard $i$ to yard $j$), the objective is to minimize the total travel time for all trains, including delays due to malfunctions. The mathematical formulation is as follows:

\begin{equation}
\begin{aligned}
    &\min \sum_{r=1}^{N_{tr}} \sum_{(i,j) \in E_r} \left( \frac{d_{ij}}{v_r} + m_r h_r \right) x_{ij}^r \\[6pt]
    &\text{subject to:}\\
    &\sum_{j \in V} x_{ij}^r = 1 
    \quad 
    \forall\, r = 1,\ldots,N_{\text{tr}}, \ \forall\, i \in V,\\[6pt]
    &\sum_{i \in V} x_{ij}^r = 1 
    \quad 
    \forall\, r = 1,\ldots,N_{\text{tr}}, \ \forall\, j \in V,\\[6pt]
    %
    &\sum_{i \in S} \sum_{j \in S} x_{ij}^r \;\le\; |S| - 1
    \quad 
    \forall\, r = 1,\ldots,N_{\text{tr}}, \ \forall\, S \subset V : 2 \le |S| \le v - 1,\\[6pt]
    &x_{ij}^r \in \{0,1\} \quad \forall\, i,j \in V, \ \forall\, r = 1,\ldots,N_{\text{tr}}.
\end{aligned}
\end{equation}

\looseness=-1

The first two constraints ensure that each train must have a complete route from a source yard to a destination yard.  The third constraint, called the sub‐tour‐elimination constraint, ensures that no proper subset of yards forms an isolated cycle, thereby preventing a train from visiting any yard more than once.

Traditionally, Operations Research (OR) techniques are employed to tackle the TSP, leveraging mathematical modeling and optimization to systematically improve train path scheduling \cite{lusby2011railway, dong2020integrated}. These methods break down the TSP into a graph optimization problem to identify non-intersecting routes for trains in time and space. While OR methods provide structured solutions, they face notable limitations, particularly in computational efficiency and adaptability. Computing exact solutions incurs high overhead, and frequent rescheduling is required in this domain to accommodate disruptions such as infrastructure failures, or train delays \cite{khadilkar2018scalable}.


\subsection{Flatland Simulator}
The Flatland Simulator is a two-dimensional simplified grid environment that allows for faster and diverse experimentation in  TSP. Flatland provides an easy-to-use interface to test novel approaches for the TSP \cite{mohanty2020flatland}. Agents are tasked with efficiently navigating trains from start to finish. The score is computed based on the successful routing of trains to their destinations and the completion of scenarios, where all trains within an environment reach their respective endpoints.


\section{Curriculum Design for the Train Scheduling Problem}

 To construct an effective curriculum, it is essential to consider task generation, sequencing, and transfer learning. A well-designed curriculum improves both accuracy and speed by breaking down the train scheduling problem into distinct subproblems enabling better performance and faster convergence.  Our approach specifically targets four key challenges essential for TSP:



\begin{itemize}
\item \textbf{Pathfinding}: An agent is expected to navigate the network efficiently from start to destination without unnecessary detours, recognizing that optimal pathfinding is foundational to handling complex scheduling scenarios.
\item \textbf{Train-Speed Differences}: An agent should also address conflicts arising from varying train speeds. Slower trains can hinder the progress of faster ones, necessitating strategic planning to mitigate such issues.

\item \textbf{Malfunctions}: An agent should be able to handle random malfunctions, necessitating adaptive responses to avoid delays.
\item \textbf{Deadlocks}: It is critical that an agent efficiently handles various types of deadlocks (common, cyclic, and track-caused). An agent should efficiently plan and avoid paths that might lead to these scenarios, even at the cost of increased travel time for some trains.

\end{itemize}

These challenges often occur in combination, requiring agents to make decisions that minimize the impact of such scenarios on overall system efficiency.  We design a curriculum tailored to address specific  problems by alternating between three specialized environment types focused on pathfinding, malfunctions and train-speed differences, and deadlocks (see Figure \ref{fig:main}).

\begin{itemize}
\item Pathfinding starts with a simple 4x4 environment focusing on single-agent navigation, gradually increasing in size to enhance the agent's goal-reaching capabilities.
\item Malfunctions and train-speed differences are taught in a unified setting due to their overlapping problem-solving strategies, emphasizing decision-making between following closely and opting for longer, less congested routes.
\item Deadlock avoidance focuses on the most common deadlock scenarios, where  two or more trains become mutually blocked at junctions or intersections and preventing any movement. 
\end{itemize}

We sequence the tasks in the following way: Pathfinding, malfunctions and train-speed adjustments, and deadlock avoidance. We posit that starting with pathfinding is the most sensible choice, as all other tasks require an efficient pathfinding strategy. We leave deadlock avoidance for last, as it requires the agent to master the skills of pathfinding and handling malfunctions and various train speeds.

Furthermore, we consider two versions of our curriculum: with and without rehearsal. Rehearsal involves revisiting earlier tasks to reinforce learning, aiming to examine its impact on knowledge retention and application across varying complexities. The curriculum's structure is designed to equally distribute learning across different environments. See the appendix for detailed parameters of the curriculum.



\begin{algorithm}[tbh]
\label{algo}
\caption{Continual DQN Expansion (CDE)}
\begin{algorithmic}[1]
\REQUIRE $\theta_1, \ldots, \theta_j$ (previous Q- Approximators), $\epsilon$ (threshold), New Task $k$,
\STATE Initialize replay buffer $R_k \leftarrow \emptyset$
\STATE Initialize: $\theta_{j+1} \leftarrow \frac{1}{j} \sum_{i=1}^{j} \theta_i$ with learnable Rational Activation Functions 
\FOR{ Episode $e = 1$ ... $E$}
    \STATE Initialize a random process $\mathcal{N}$ for action exploration
    \STATE Receive initial observation state $s_1 \sim p_{K}(s_1)$ for the new task $k$. 
    \FOR{$t = 1$ ... $T_k$}
       \STATE Select a random index $i \in \{0, \ldots, j+1\}$
\STATE Select action \( a_t = \mu(s_t|\theta^{\mu}_{i}) + \mathcal{N}_t \)

    \STATE Execute \( a_t \) and observe \( r_t \) and \( s_{t+1} \)
    \STATE Store transition \( (s_t, a_t, r_t, s_{t+1}) \) in \( R_k \)
    \ENDFOR
    \IF{Weight Update Step}
        \STATE Update The Q-Network with weights $\theta_{j+1}$ and their adaptive activation functions. 
        \STATE Update The previous Q-Networks $\theta_1, \ldots, \theta_j$ with Loss  
        $L(\theta_i) = L(\theta_i) + 
        \sum_{i} \frac{\lambda}{2} F_{i} (\theta_{i} - \theta_{i}^{*})^2$ where  \( \theta_{i}^{*} \) are the weights of the models trained on the previous task.
    \ENDIF
\ENDFOR
\IF{$W_\phi(\theta_{\text{j+1}}) > \epsilon \cdot W_\phi(\theta_{i})$ \text{ for all } $i \in \{1, \ldots, j\}$}
    \STATE \text{find } $k = \arg\min_{i \in \{1, \ldots, j\}} W_\phi(\theta_{i})$
    \STATE return $\theta_1, \ldots, \theta_{k-1}, \theta_{k+1}, \ldots, \theta_j, \theta_{j+1}$;
\ELSE
    \STATE return $\theta_1, \ldots, \theta_j$;
\ENDIF

\end{algorithmic}
\label{algo}
\end{algorithm}

\section{Continual DQN Expansion Algorithm}
 We propose a novel approach for adaptively constructing Q-function subspaces to address the challenges of non-stationary environments. Our Continual  Deep Q-Network (DQN) Expansion (CDE) approach dynamically generates and adjusts Q- function subspaces, denoted as $\Theta_1, \ldots, \Theta_N$, based on environmental changes and task requirements.  By selectively expanding these subspaces and employing elastic weight consolidation (EWC), our method mitigates catastrophic forgetting, while a rational adaptive activation function enhances plasticity for rapid task adaptation. In the following sections, we review EWC and Padé activation units (PAU) and describe how our methodology integrates subspace expansion with PAU and EWC.


\subsection{Formalisation of the Continual Reinforcement Learning Problem}
\looseness=-1

A continual reinforcement learning scenario is characterized by a sequence of $K$ tasks or instances. Each task $k$ is described through a Markov Decision Process (MDP) $M_k = \langle S_k, A_k, T_k, r_k, \gamma \rangle$, which encompasses a set of states $S_k$, a set of actions $A_k$, a transition function $T_k: S_k \times A_k \rightarrow \mathcal{P}(S_k)$, and a reward function $r_k: S_k \times A_k \rightarrow \mathbb{R}$. We assume in our case that all tasks share identical state and action spaces.

\looseness=-1

A global Q-function $Q: [1..K] \times \mathcal{Z} \rightarrow (S \times A \rightarrow \mathbb{R})$ is defined. This function accepts a task identifier $k$ and a sequence of tasks $\mathcal{Z}$ as inputs, yielding a Q-value representing the expected future rewards for interacting with the environment of $k^{\text{th}}$ task. The output is a real-valued estimate of the expected return from taking action $A$ in state $S$.

\looseness=-1

For each new task, we reset the replay buffer. Meaning that upon transitioning to task $k+1$, the transitions from the previous MDP $M_i$ are no longer accessible  to the agent.

\subsection{Elastic Weight Consolidation}
 The main idea  behind EWC  is that given a sequence of tasks, after learning task $k$, the parameters $\theta$ of the neural network are adjusted not only based on the current task's loss but also considering the importance of these parameters to the previously learned tasks $1, \ldots, k-1$.  EWC introduces a regularization term to the loss function \cite{kirkpatrick2017overcoming}. The regularization term  is  computed based on the Fisher information matrix $F$ from the previous tasks, which estimates the importance of each parameter to the tasks already learned.
This term effectively constrains the parameter updates, favoring updates that have minimal impact on the performance of past tasks.  The loss function $L(\theta)$ that we minimize in EWC is defined as:
 
\begin{equation}
\label{eq:ewc-loss}
L(\theta) \;=\; L_k(\theta) 
\;+\; \sum_{j=1}^{k-1} \frac{\lambda}{2} \sum_{i} F_{j,i}\, \bigl(\theta_i - \theta_{j,i}^*\bigr)^2,
\end{equation}

where $L_k(\theta)$ is the loss function for the current (i.e., $k$-th) task,
$\lambda$ issets the importance of the old task relative to the new one. $F_{j,i}$ is the Fisher information for parameter $i$ from task $j$ and 
 $\theta_{j,i}^*$ denotes the value of the parameter $\theta_i$ after training on task $j$.


When moving on to the next task, EWC aims to keep the network parameters close to those learned on previous tasks. This can be achieved either through multiple separate penalty terms or a single combined penalty term, since the sum of multiple quadratic penalties is itself also quadratic.

\subsection{Pad\'e Activation Units}
Pad\'e activation units (PAU) enhance the adaptability and efficacy of neural networks in continual reinforcement learning environments \cite{delfosse2021adaptive}. PAUs, defined as the ratio of two polynomial functions, offer a flexible and learnable activation mechanism that can dynamically adjust to varying task complexities. This adaptability is particularly beneficial in continual RL settings, where the agent encounters a sequence of tasks, each potentially differing in its dynamics and objectives.

In a continual RL setting, employing PAUs allows for the modulation of the activation functions based on the specific requirements of each task, thus improving the plasticity of neural networks. Formally, PAU is defined as: 
\begin{align}
    R(x) = \frac{\sum_{j=0}^{m} a_j x^j}{1 + \sum_{k=1}^{n} b_k x^k}
\end{align}
 where $a_j$ and $b_k$ are the learnable parameters that determine the shape of the activation function. The adaptability of PAUs lies in their ability to represent a wide range of nonlinear relationships through the adjustment of these parameters, thereby enabling a neural network to maintain robust performance across a continuum of RL tasks and non-stationary environments.

\subsection{Subspace Expansion of Q-Functions}

We introduce the Continual Deep Q-Network (DQN) Expansion (CDE) approach, a novel method designed to adaptively construct Q-function subspaces for continual reinforcement learning. Our approach facilitates sequential task learning by an agent. The pseudo-code is provided in Algorithm \ref{algo}.

The CDE algorithm generates a sequence of subspaces, denoted as $\Theta_1, \ldots, \Theta_N$, each corresponding to a new change in the agent's environment. The dynamic expansion process involves the integration of new anchors $\theta$ function, defined by a neural network. A new anchor is added only if it demonstrates a superior performance  over the previous anchors within the existing subspace, ensuring sublinear expansion of the number of anchors. Our expansion is limited to exactly N networks as we remove the worst-performing anchors each time we add new ones. N is a hyperparamter that depends on the continual learning task. In our main results, we set $N=2$. We further analyze expansion size and timing (subtask vs main-task expansion) in the ablation study.

CDE maintains the current anchors set composed of the weight vectors $\{ \theta_1, \ldots, \theta_N \}$ that delineate the optimal Q-functions for previously encountered tasks. Upon the introduction of a new task or subtask, denoted as $k+1$, the models creates a preliminary new subspace $\Theta_{j+1}$. We initialize the weights  of the new Q-function network as follows:
\[
    \theta_{j+1}=\frac{1}{j} \sum_{i=1}^{j} \omega_i \cdot \theta_i
\]
where $\omega_i$ is the coefficient of each previous anchor. 
To mitigate catastrophic forgetting of previous tasks, the weights $\{ \theta_1, \ldots, \theta_N \}$  of the previous subspaces are refined using an EWC  regularized loss. The anchor $\theta_{j+1}$ for the new subspace is optimized with rational adaptive activations to enhance plasticity and accelerate the adaptation process.

The decision-making process within CDE involves either identifying an effective pre-existing Q-function for the new task within the current subspace or creating a new subspace with a more adequate anchor $\theta_{j+1}$ for the new task. If the new subspace achieves higher performance and we have already reached the maximum subspace size, we remove the worst-performing subspace in the pruning step.


\looseness=-1
Our approach can be summarized in three main steps: (1) expanding the current subspace by introducing a new anchor, (2) training and adjusting all anchors for the current task, and (3) evaluating the efficacy of the new anchor relative to the optimal anchor function within the existing subspace.

Based on this comparison, CDE determines whether the subspace should be further extended (i.e., adding a new anchor to the subspace set) or reverted to its previous configuration (keeping the subspace set as it is) or pruned (adding a new anchor and removing the worst-performing anchor). This ensures a balanced adaptation to new tasks while preserving previously acquired relevant knowledge.

\section{Experiments}


\subsection{Training Curricula}
 The standard approach to training TSP reinforcement learning agents is to train them directly in the target environment without preliminary steps. For Flatland, this involves generating 64×64 environment with 14 agents using a sparse line generator. However, the large search space leads to worse generalization and slower convergence. In this work, we explore the interplay between algorithmic design and curriculum design to enhance generalization for TSP.  We train   various reinforcement learning algorithms under the following learning conditions: (1) a baseline with no curriculum (i.e., a single 64×64 environment with 14 agents), (2) dynamic or naive environment curriculum (i.e., a sequence of environments with increasing sizes from 16×16 to 64×64 and a corresponding increase in the number of agents from 1 to 14), and (3) our proposed custom curriculum , which not only scales the environment and agent numbers but also introduces increased task complexity as illustrated in Figure \ref{fig:main}.  Further details are in Table \ref{table:curriculum}. 

The dynamic environment curriculum begins with less complex environments and progressively increases in difficulty by enlarging the environment size and agent count. This gradual introduction of complexity helps agents adapt to more challenging scenarios. Our proposed custom curriculum addresses specific TSP challenges by alternating between three specialized environment types focused on pathfinding, malfunctions and train-speed differences, and deadlocks.

\subsection{Experimental Setting}

To evaluate the performance of different algorithms, we utilize a simulated environment that replicates a realistic and complex railway system \cite{leichthammer2022evaluating}. Each algorithm is tested without undergoing a prior adaptation phase on the test environment.
We use both score and completion rate as evaluation metrics. The score also serves as the reward signal used to train the  agents and is computed based on the distance of each train to its destination. A lower score indicates better performance, as it means trains are closer to their targets.
The completion rate measures the percentage of trains that successfully reach their destination. Notably, a lower score does not always imply a higher completion rate as agents may significantly reduce their distances without fully completing their routes. Furthermore, small improvements in score often require substantial gains in completion rate. 

\subsection{Baselines and Hyperparamters}

 Our baselines consist of an Advantage Actor-Critic (A2C), Proximal Policy Optimization (PPO) , and Deep Q-Network (DQN) algorithm \cite{schulman2017proximal, mnih2015human}. Each algorithm utilizes two hidden layers with 1024 units each, a replay buffer of size 1 million a batch size of $128$, and a learning rate of $ 5 \cdot 10^{-5}$. We use a simulated annealing exploration strategy. In addition, we include PackNet \cite{mallya2018packnet} and BFF \cite{schwarzer2023bigger} baselines, following their original implementations.

For PAUs, we set \( m = 5 \) and \( n = 4 \), initialize them with the shape of a standard RELU. For EWC, we use \( \lambda = 0.5 \). During the expansion step of CDE, new anchors are initialized with binary coefficients \( \omega_i \), meaning we copy the weights of the most recent anchors as a starting point. Further implementation details are included in the appendix. 

\subsection{Results and Discussion}

\vspace{-0.38cm}
\begin{table*}[ht]
\caption{Performance comparison of different baselines trained with a range of curricula. The completion rate (higher is better $\uparrow$) is reported. An extended version of the table with the scores is found in the  The best score and completion rate are highlighted in bold. We see that the custom curriculum with decomposed tasks (see Figure \ref{fig:main}) achieves the highest performance. Further improvements are observed for the custom curriculum  with rehearsal in the case DQN. }

\centering
\begin{tabular}{lcccc}
\toprule
\textbf{Method} & \textbf{No Curric. } & \textbf{Naive Curric.} &  \textbf{Custom Curric.}  & \textbf{Custom Curric. (Rehear.)} \\

\midrule
A2C    & $0.06 \pm 0.01$ & $0.07 \pm 0.03$ & $0.05 \pm  0.02$   & $0.06 \pm  0.02$  \\
PPO & $0.06 \pm 0.02$  &  $0.06 \pm 0.02$  & $0.06 \pm  0.02$   & $0.06 \pm  0.03$  \\

DQN   & $0.04 \pm 0.07$ &  $0.08 \pm 0.04$ & $0.10 \pm  0.04$   & $\mathbf{0.19 \pm  0.04}$ \\

\bottomrule
\end{tabular}
\label{table:curriculum_results}

\end{table*}
\looseness=-1

Table \ref{table:curriculum_results} compares the performance of standard RL algorithms under different curricula on the evaluation environment. Incorporating a curriculum consistently improves scores and completions for all baselines. For instance, DQN achieves a score of $0.04 \pm 0.07$ without a curriculum, compared to $0.19 \pm 0.09$ with a curriculum. Notably, our proposed curriculum yields greater improvements in DQN than a naive curriculum, highlighting the benefits of a more structured approach. However, for PPO and A2C, the gains remain within the margin of error, suggesting that curriculum design should align closely with the underlying algorithm.

\looseness=-1
Furthermore, we observe that rehearsal significantly improves DQN's performance, nearly doubling the completion rate from $0.10$ with the proposed custom curriculum to $0.19$ with the proposed custom curriculum plus rehearsal. This underscores the need to address catastrophic forgetting to learn a more generalizable strategy, which we aim to achieve with our proposed method, CDE. In the proposed curriculum, specific skills are learned in each new environment, so it is crucial to ensure the agent retains previous skills. For example, pathfinding skills are essential for handling varying train speeds, illustrating how tasks build on each other. 

\looseness=-1
In Table \ref{table:cde}, we extend DQN with different techniques and compare the generalization performance when training using our custom curricula. We observe that increasing plasticity (DQN +PAU) or mitigating catastrophic forgetting (DQN +EWC) helps achieve more generalizable minima compared to a simple DQN. However, both methods perform worse than simple rehearsal ($0.19$ completion for DQN +EWC and $0.12$ for DQN +PAU compared to $0.19$ for DQN with rehearsal). Additionally, in this scenario, addressing catastrophic forgetting is more beneficial than enhancing plasticity. This is because rational activations are most advantageous when the neural network begins to lose plasticity after a large number of tasks (e.g., more than 10 tasks). In contrast, our curriculum involves only three tasks, each comprising four different environment sizes.

\looseness=-1
A naive combination of both approaches for plasticity (PAU) and stability (EWC) results in performance degradation for DQN +EWC+PAU compared to just DQN +EWC. This can be explained by the opposing effects of rational activations and elastic weight consolidation, where changing activations render the Fisher matrix weights for EWC  obsolete.

\begin{table}[ht]
\caption{Performance comparison of different algorithms trained using our custom curriculum without rehearsal. The score (lower is better $\downarrow$) and completion rate (higher is better $\uparrow$) are reported. The best score and completion rate are highlighted in bold. We see that our proposed method CDE achieves the highest performance and benefits the most from the custom curriculum.}
\centering
\begin{tabular}{lcc}
\toprule
\textbf{Method} & \textbf{Score} & \textbf{Completion Rate}\\
\midrule
DQN        & 0.34 ± 0.01 & 0.10 ± 0.04 \\
DQN w/ Rehear.   & 0.31 ± 0.02 & 0.19 ± 0.04 \\
\midrule
DQN +EWC \cite{kirkpatrick2017overcoming}    & 0.31 ± 0.01 & 0.19 ± 0.08 \\
DQN +MAS \cite{aljundi2018memory}    & 0.33 ± 0.01 & 0.13 ± 0.03 \\

\midrule
DQN +PAU \cite{delfosse2021adaptive}  & 0.32 ± 0.01 & 0.12 ± 0.04   \\
DQN +CBP \cite{dohare2021continual}  &  0.34 ± 0.01  &  0.09 ± 0.05 \\ 
\midrule
PackNet \cite{mallya2018packnet}   & 0.34 ± 0.01 & 0.16 ± 0.05 \\
BFF \cite{schwarzer2023bigger}   & 0.31 ± 0.01 & 0.19 ± 0.04 \\
\midrule

DQN +PAU+EWC   & 0.32 ± 0.01 & 0.18 ± 0.05 \\
CDE (ours) & \textbf{0.29 ± 0.01} & \textbf{0.28 ± 0.06}\\
\bottomrule
\end{tabular}

\label{table:cde}
\end{table}
\looseness=-1
Instead, our method (CDE) ensures the network remains both plastic and stable depending on the current task. CDE achieves the best performance, with almost a 50\% improvement in the completion rate compared to DQN with rehearsal, even without using curriculum rehearsal during training of CDE. Our approach balances plasticity and stability by expanding the network in a targeted manner based on the current task and the performance of pre-existing subspaces.  Further results in the appendix show that CDE switches between more stable and more plastic networks depending on task requirements and is less affected by task ordering. Unlike the baselines, which tends to overfit early tasks like pathfinding and suffers from catastrophic forgetting in later ones, CDE gradually builds transferable skills across tasks and maintains high performance on the generalization environment regardless of the task sequence see Figure \ref{fig:Dqn} and \ref{fig:CDE} in the appendix.

We refine network expansion in CDE by comparing two strategies: expanding with every environmental change (e.g., adding agents or increasing size) versus expanding only for distinct tasks as presented in Figure \ref{fig:main}. The former corresponds to subtask expansion, where the underlying task remains the same but complexity increases (e.g., larger environments or more agents), while the latter involves expansion between tasks (e.g Patfinding and Malfunctions), where the core challenges in the environment differ.

\begin{figure*}[h]

    \centering
    \subfigure[]{\includegraphics[height=4.5cm]{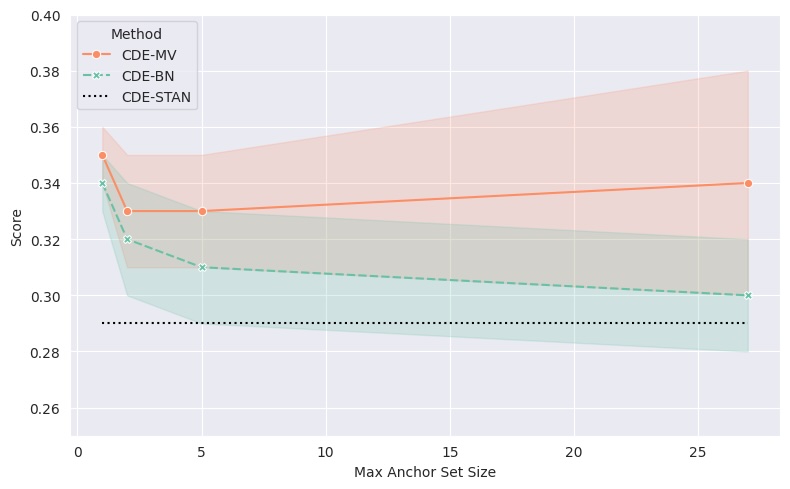}}
    \hspace{0.01cm}
    \subfigure[]{\includegraphics[height=4.5cm]{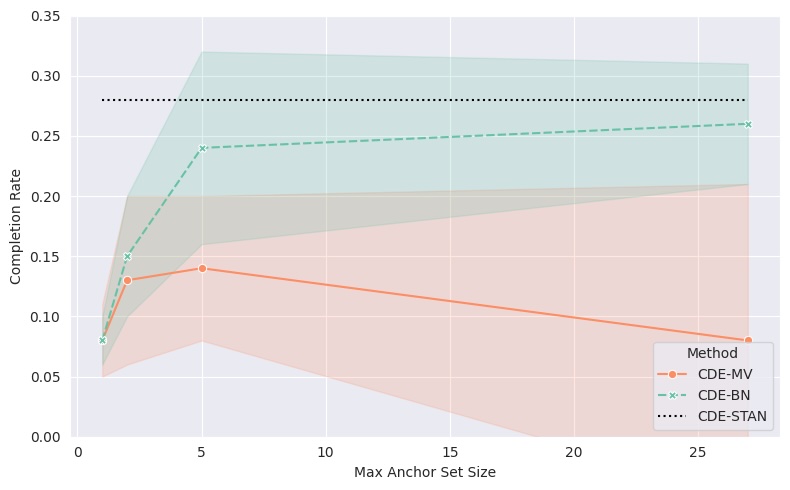}}
    \hspace{0.01cm}
    \vspace{-0.3cm}
    \caption{Comparison of performance metrics for different CDE expansion strategies. The plots show the mean score (left) and completion rate (right) as a function of the number of expansions. The methods are abbreviated as follows: \textbf{CDE-BN} (subtask expansion with best network selection), \textbf{CDE-MV} (subtask expansion with majority vote), and \textbf{CDE-STAN} (standard task expansion). }
    \label{fig:cde_plots}
\end{figure*}

In Figure \ref{fig:cde_plots},  we observe that expansions triggered by changes in the size of the environment do not consistently increase performance and often increase variance. Although increasing the size of the anchor set eventually recovers performance comparable to the original CDE, it also raises computational overhead due to a larger number of anchors to update. This indicates that catastrophic forgetting is less problematic when tasks change minimally, and expansions offer greater benefits when tasks shift substantially (e.g., from pathfinding to managing malfunctions).




\textbf{Limitations and Future Works.} Despite its strong empirical performance, the CDE framework has several notable limitations. First, the use of multiple subspaces introduces significant computational overhead. Even with subspace limiting and network rotation, training time remains substantially higher (Table \ref{table:training_time}), and this cost will furhter increase as curricula become longer. Second, our experiments have been conducted on relatively short, fixed curricula under a single form of non‐stationarity; it remains an open question for future work how CDE’s design principles will scale to much longer training schedules or to environments with more dynamic task shifts (e.g., diverse Atari benchmarks). Finally, although CDE narrows the gap, it still lags behind classical operations‐research methods on the train‐scheduling problem, underscoring the challenge of combining the adaptability of RL based approaches with the optimality guarantees of traditional algorithms.

\begin{table}[ht]
\centering
\caption{Training time of each method in wall-clock time (seconds). Here we only count the effective training time, not the environment generation and loading in the Flatland simulator. The total number of iterations is fixed across methods. We observe that the increased complexity of our method CDE leads to longer training time for the same number of iterations, due to updating multiple anchors at each iteration.}
\begin{tabular}{lc}
\toprule
\textbf{Method} & \textbf{Training Time (s)} \\
\midrule
A2C & \: 889 ± \: 64 \\
PPO & \: 812 ± \:  72 \\
DQN                           & 1182 ± \: 84   \\
\hline
DQN +EWC \cite{kirkpatrick2017overcoming}   & 1206 ± \: 80   \\
DQN +MAS \cite{aljundi2018memory}           & 1198 ± \: 76   \\
\hline

DQN +PAU \cite{delfosse2021adaptive}        & 2128 ± 102  \\
DQN +CBP \cite{dohare2021continual}         & 1655 ± \: 96   \\
\hline

DQN +PAU+EWC                               & 2178 ± \: 89   \\
CDE (ours)                                 & 3138 ± 138  \\
\bottomrule
\end{tabular}

\label{table:training_time}
\end{table}

\section{Conclusion}

\looseness=-1
We presented a curriculum for TSP and a continual expansion approach to further improve the performance. Our curriculum design philosophy emphasizes a comprehensive approach to agent training, addressing key skills to enhance generalization to more complex environments. Our results show that curriculum design should transcend increasing the size of the environment and instead be tailored to the specific application domain. Fully leveraging such curricula requires a careful balance between retaining previous skills and acquiring new ones, ensuring that agents can adapt to dynamic challenges without falling into suboptimal policies. Our Continual DQN Expansion approach effectively manages the interplay between plasticity and stability. By demonstrating superior improvements and minimal forgetting, our method addresses the limitations of previous approaches. Our results highlight the importance of a holistic curriculum and algorithm design for improved generalization. Thus, there is a need for ongoing research to scale these approaches across different domains for more robust and domain specific continual learning agents.

\section*{Acknowledgments}
This work was supported by the KIBA Project (Künstliche Intelligenz und diskrete Beladeoptimierungsmodelle zur Auslastungssteigerung im Kombinierten Verkehr - KIBA) under reference number 45KI16E051, funded by the German Ministry of Transportation.

\bibliography{collas2025_conference}
\bibliographystyle{collas2025_conference}
\appendix
\section{Appendix: Mitigating the Stability-Plasticity Dilemma in Adaptive Train Scheduling with Curriculum-Driven Continual DQN Expansion}

In the supplementary material, we provide additional details, experimental setup and descriptions for the various parts  of our proposed frameworks as well as discuss some additional results. The structure is as follows:

\begin{itemize}
    \item Implementation Details for Continual DQN Expansion Algorithm (CDE)
    \item Hyperparameter Selection Process
    \item Compute Resources
    \item Curriculum Details
    \item Additional Experiments
    \item Limitations and Future Works

    \item Reproducibility checklist

\end{itemize}

\section*{Continual DQN Expansion (CDE) Algorithm}

The Continual DQN Expansion (CDE) algorithm manages and expands a set of subspaces defined by their anchors, which are Q-Function networks that are adapted to new tasks while retaining knowledge from previous ones. Below is a detailed breakdown of the CDE process:

\subsection*{Subspace Set Management}

The subspace set is implemented in PyTorch as a list of lists, where each sublist contains Q-networks. These networks are dynamically managed and expanded as learning progresses. During training, actions are selected using the network indicated by the \texttt{act\_rotation} index, which cyclically rotates among the available networks. After each action, the current network’s performance (score and completion) is updated, and the rotation index (\texttt{act\_rotation}) is incremented, allowing for the cyclical usage of multiple networks.

\subsection*{Learning Process}

\begin{figure*}[h]
    \centering
    
    \includegraphics[width=0.3\textwidth]{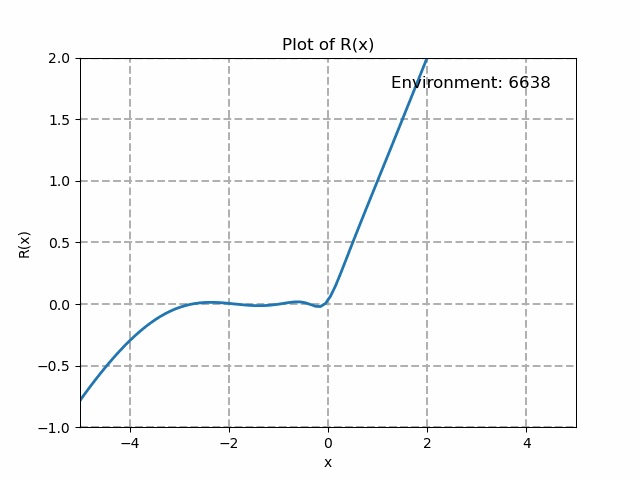}
    \hfill
    \includegraphics[width=0.3\textwidth]{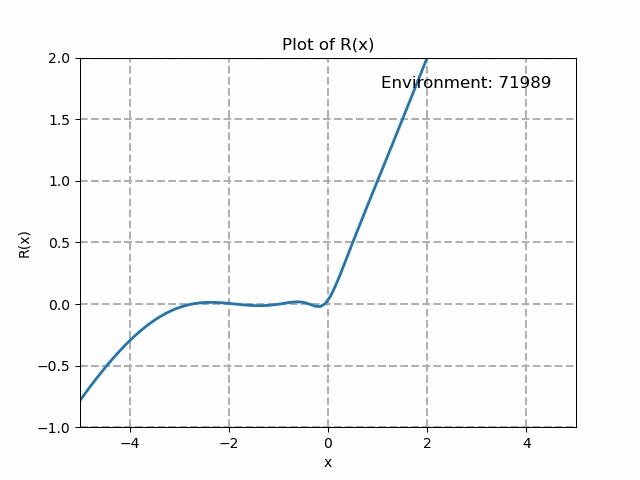}
    \hfill
    \includegraphics[width=0.3\textwidth]{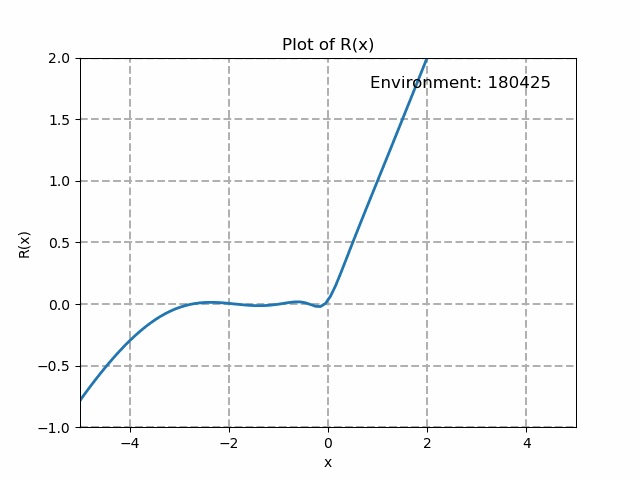}
    
    
    \includegraphics[width=0.3\textwidth]{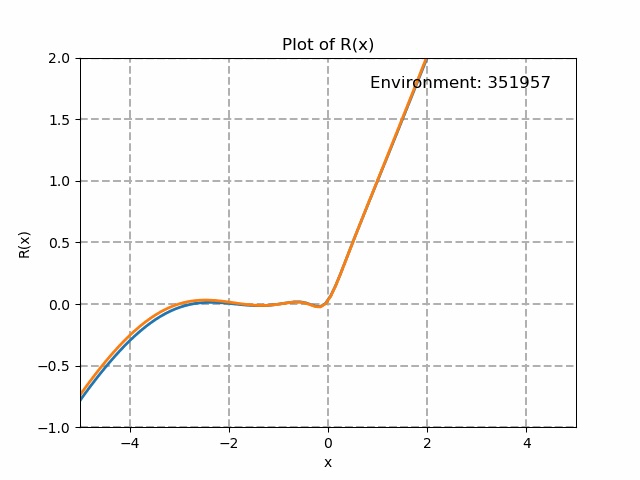}
    \hfill
    \includegraphics[width=0.3\textwidth]{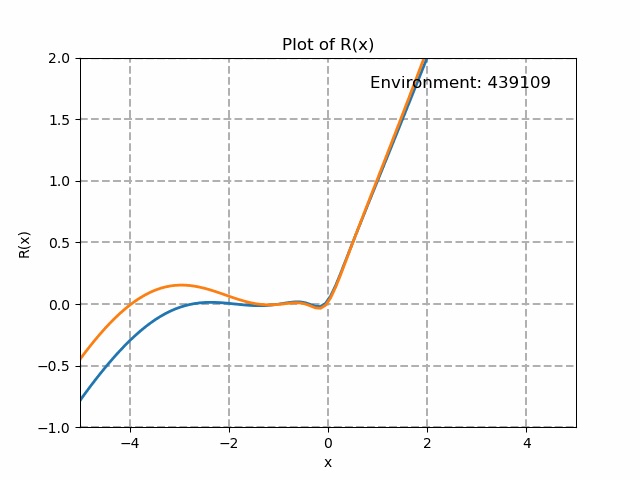}
    \hfill
    \includegraphics[width=0.3\textwidth]{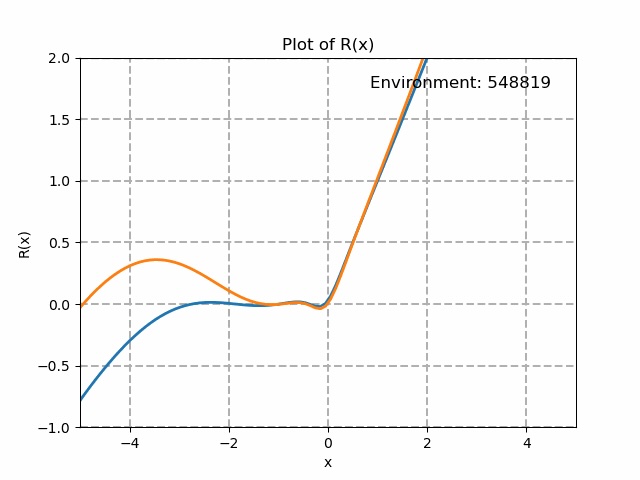}
    
    
    \includegraphics[width=0.3\textwidth]{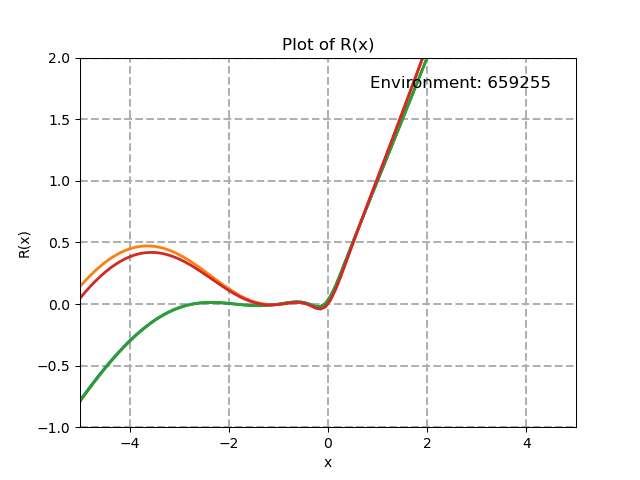}
    \hfill
    \includegraphics[width=0.3\textwidth]{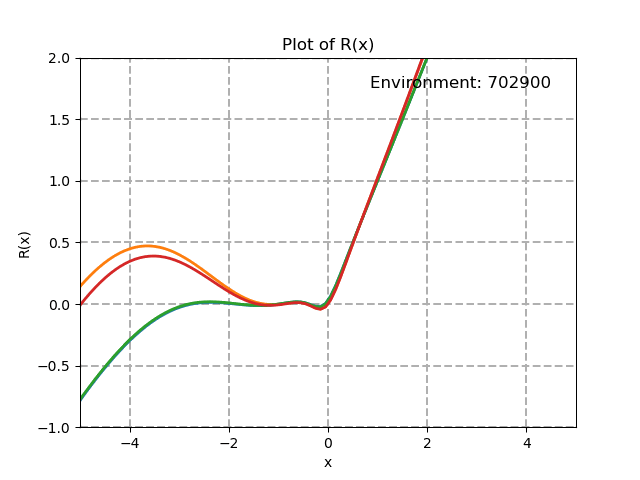}
    \hfill
    \includegraphics[width=0.3\textwidth]{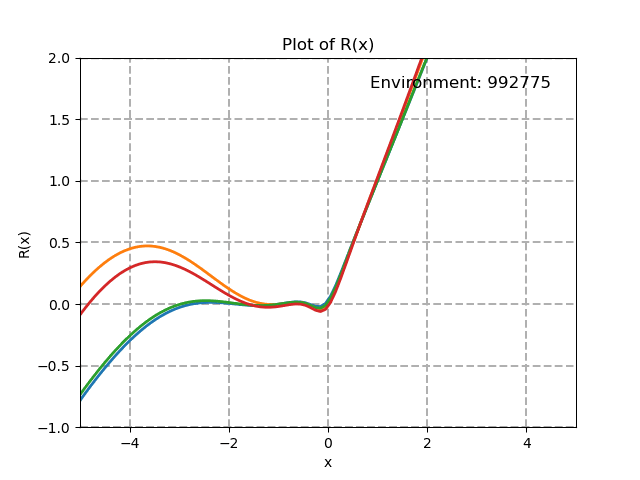}
    
    \caption{Evolution of learned activation functions for CDE. The plots show how newer anchors are added as the number of tasks increases, with each developing distinct activation units indicated by the different colors. Over time, some anchors maintain fixed activations, while others continue to evolve with newer tasks (orange vs. red, or green vs. blue). This demonstrates how CDE effectively creates varied subspaces, represented by anchors in the form of neural networks.}
    \label{fig:PAU_Activations}
\end{figure*}

Experience replay is utilized during training, ensuring that each network in the current list is updated with the latest experiences. The \texttt{\_learn} method performs Q-learning updates and can incorporate Elastic Weight Consolidation (EWC) regularization to safeguard against catastrophic forgetting, preserving knowledge from previously established anchors. In contrast, newly initialized anchors are equipped with adaptive rational activation functions, enhancing their plasticity and enabling faster adaptation to new tasks. See Figure \ref{fig:PAU_Activations} for a visualization of the evolution of the activation functions of various anchors as the training progresses.

\subsection*{Network Expansion Strategies}

A novel aspect of CDE is its ability to expand the set of networks when new tasks are encountered, coupled with the continuous adaptation of existing anchors through regularization methods. During the expansion process, two key strategies are employed:

\begin{enumerate}
    \item \textbf{Elastic Weight Consolidation (EWC):}
    
    Previous anchors are trained using EWC, which helps preserve important knowledge from previous tasks by penalizing significant changes to parameters deemed important (as identified by the Fisher Information Matrix).

    \item \textbf{Pade Activation Units (PAU):}
    
    Another version of the network, which denotes the anchors of the new subspace, is trained using Pade Activation Units (PAU). PAUs offer a more flexible and expressive form of activation function, allowing the network to better adapt to complex tasks and maintain high plasticity during later tasks.
\end{enumerate}

During expansion, the algorithm evaluates the performance of the current networks and retains the top performers. The expansion process adds new networks to the set, initialized with parameters derived from the best-performing networks, and trains them using both EWC and PAU. When in evaluation mode, the system selects the best-performing network based on the average score. This network is then used to determine the action.

Below is a high level pseudocode for  the CDE implementation in Pytorch:
\begin{lstlisting}[basicstyle=\small\ttfamily, breaklines=true, frame=none, backgroundcolor=\color{white}]
Continual_DQN_Expansion:
    Initialize:
        - state_size, action_size, parameters, evaluation_mode
        - networks = [[]]
        - Add initial policy network to networks[0]
        - Initialize various variables: act_rotation, networkEP, networkEP_scores, networkEP_completions

    act(handle, state, epsilon=0, eval=False):
        if eval:
            - Select the network with the best average score
            - Return action from the selected network
        else:
            - Return action from the current network in rotation

    network_rotation(score, completions):
        - Update the score (networkEP_scores) and completions (networkEP_completions) for the current network
        - Increment act_rotation to the next network

    step(handle, state, action, reward, next_state, done):
        - Update all networks in the current list with the new experience. 

    expansion_and_pruning():
        - Prune: Select top n networks based on a defined score metric.
        - Update the network list with the selected networks.
        - Initialize the Fisher Matrix for EWC updates for each network.
        - Freeze adaptive activations for each network in the list.
        - Expand: Add new networks with adaptive activation functions.
        - Adjust act_rotation accordingly.
\end{lstlisting}

\section*{Hyperparamters Selection}
In this work, we selected the hyperparameters for the different algorithms to ensure optimal performance, reproducible results and stable convergence. 
These values were chosen based on hyperparameter sweeps provided by the Flatland Community \footnote{\url{https://wandb.ai/masterscrat/flatland-examples-reinforcement_learning/reports/Flatland-Starter-Kit-Training-in-environments-of-various-sizes--VmlldzoxNjgxMTk}}. We use the same hyperparameters for our CDE algorithm and other extensions mentioned in the main body.

The learning rate was set to 0.00005. The discount factor (gamma) was set to 0.99. For exploration, the epsilon-greedy strategy was employed, with the initial exploration rate (\(\epsilon\)) set to 1.0 to ensure sufficient exploration in the early stages of training. The epsilon value was then decayed according to the formula \(\epsilon(x) = \frac{1}{1.000005^x}\), where \(x\) represents the number of steps taken. This exponential decay gradually reduces epsilon, allowing the agent to shift its focus from exploration to exploitation as it gains more experience. 

The weights of the Q-network were updated every 8 steps with batch of size $128$. Additionally, the replay buffer size was set to 1000,000. We note that we empty the replay buffer when we change the task, ensuring that we only use experience from the current task.

\section*{Compute Resources}
Each algorithm was trained using one Intel Xeon Gold 5218 CPU @ 2.3Ghz and one NVIDIA V100  SXM2 32GB GPU. Each run took between approximately 3 and 10 hours to complete, dependent on the algorithm and Curriculum used. We repeat each of the experiments 10 times.
\section*{Curriculum Details}

Table \ref{table:curriculum}  presents details for the various curriculum strategies applied used in the main body. The experiments are divided into four categories: \textbf{No Curriculum}, \textbf{Simple Curriculum}, \textbf{Custom Curriculum without Rehearsal}, and \textbf{Custom Curriculum with Rehearsal}. Each category outlines specific configurations for the environment, including the environment size, the number of agents, the training speed, malfunction rates, and the total number of network steps.

In the \textbf{No Curriculum} setting, a sparse-generator environment of size 64x64 with 14 agents was tested, with a consistent training speed across all agents, a malfunction rate of 1 per 1000 steps, and a total of 960,000 network steps.

The \textbf{Simple Curriculum} approach gradually increases the complexity of the environment, starting from a 16x18 environment with 1 agent and ending with the same 64x64 environment used in the No Curriculum setting. This approach shows a progressive increase in the number of agents and the complexity of the tasks, with consistent training speeds and malfunction rates.
\\

The \textbf{Custom Curriculum without Rehearsal} involves different environment types, such as Pathfinding, Malfunction, and Deadlock, each tested with varying sizes and agent counts. In this strategy, the environment complexity and agent number increase, with fixed training speeds and malfunction rates. The network steps in these scenarios are capped at 80,000.

Lastly, the \textbf{Custom Curriculum with Rehearsal} revisits earlier stages in the curriculum, allowing the model to reinforce previously learned tasks while introducing new challenges. The setup is similar to the Custom Curriculum without Rehearsal but with slightly reduced network steps (64,000) to account for the rehearsal process.

This comparison allows for an assessment of how different curriculum learning strategies impact the efficiency and effectiveness of the training process across various scenarios, highlighting the benefits of gradual complexity increase and the potential advantages of rehearsal in maintaining performance on previously learned tasks.

\begin{table*}[htp]
\caption{ Overview of parameter values for each considered curriculum .}
\centering
\begin{tabular}{|p{3cm}|p{2cm}|p{1cm}|p{2cm}|p{2cm}|p{2cm}|}
 \hline
 \multicolumn{6}{|c|}{No Curriculum} \\
 \hline
 Env & Size & Agents & TrainSpeed & Malfunction & Networksteps\\
 \hline
 sparse-generator & 64x64   & 14 & 25/25/25/25& 1/1000  & 960.000\\
 \hline
 \hline
 \multicolumn{6}{|c|}{Simple Curriculum} \\
 \hline
 sparse-generator & 16x16   & 1 & 00/0/0/100& 0  & 192.000\\
 sparse-generator & 28x28   & 3 & 50/00/0/50& 1/1000  & 192.000\\
 sparse-generator & 40x40   & 6 & 50/00/25/25& 1/1000  & 192.000\\
 sparse-generator & 52x52 & 10 & 25/25/25/25& 1/1000  & 192.000\\
 sparse-generator & 64x64   & 14 & 25/25/25/25& 1/1000  & 192.000\\
 \hline
 \hline
 \multicolumn{6}{|c|}{Custom Curriculum without Rehearsal} \\
 \hline
 Pathfinding      & 4x4     & 1 & 00/00/00/100  & 0      & 80.000\\
 Pathfinding      & 8x8     & 1 & 00/00/00/100  & 0      & 80.000\\
 Pathfinding      & 16x16   & 1 & 00/00/00/100  & 0      & 80.000\\
 Pathfinding      & 32x32   & 1 & 00/00/00/100  & 0      & 80.000\\
   \hline
 Malfunction      & 12x5     & 5 & 00/00/00/100 & 1/100      & 80.000\\
 Malfunction      & 12x5     & 6 & 50/00/00/50  & 1/100       & 80.000\\
 Malfunction      & 12x5     & 7 & 50/00/25/25 & 1/100       & 80.000\\
 Malfunction      & 12x5     & 8 & 25/25/25/25& 1/100       & 80.000\\
   \hline
 Deadlock         & 32x2     & 2 & 00/00/00/100  & 1/64      & 80.000\\
 Deadlock         & 32x2     & 4 & 50/00/00/50  & 1/64      & 80.000\\
 Deadlock         & 32x4     & 8 & 50/00/25/25  & 1/64      & 80.000\\
 Deadlock         & 32x4    & 16 & 25/25/25/25  & 1/64      & 80.000\\
 \hline
 \hline
 \multicolumn{6}{|c|}{Custom Curriculum with Rehearsal} \\
 \hline
 Pathfinding      & 4x4     & 1 & 00/00/00/100  & 0      & 64.000\\
 Pathfinding      & 8x8     & 1 & 00/00/00/100  & 0      & 64.000\\
 Pathfinding      & 16x16   & 1 & 00/00/00/100  & 0      & 64.000\\
 Pathfinding      & 32x32   & 1 & 00/00/00/100  & 0      & 64.000\\
   \hline
 Malfunction      & 12x5     & 5 & 00/00/00/100 & 1/100      & 64.000\\
 Malfunction      & 12x5     & 6 & 50/00/00/50  & 1/100       & 64.000\\
 Malfunction      & 12x5     & 7 & 50/00/25/25 & 1/100       & 64.000\\
 Malfunction      & 12x5     & 8 & 25/25/25/25& 1/100       & 64.000\\
 Pathfinding      & 32x32   & 1 & 00/00/00/100  & 0      & 64.000\\
   \hline
 Deadlock         & 32x2     & 2 & 00/00/00/100  & 1/64      & 64.000\\
 Deadlock         & 32x2     & 4 & 50/00/00/50  & 1/64      & 64.000\\
 Deadlock         & 32x4     & 8 & 50/00/25/25  & 1/64      & 64.000\\
 Deadlock         & 32x4    & 16 & 25/25/25/25  & 1/64      & 64.000\\
 Pathfinding      & 32x32   & 1 & 00/00/00/100  & 0        & 64.000\\
 Malfunction      & 12x5     & 8 & 25/25/25/25& 1/100       & 64.000\\
 \hline
\end{tabular}

\label{table:curriculum}
\end{table*}

\newpage
\section*{Additional Experiments}
\begin{figure*}[h]
    \centering
    \includegraphics[width=0.49\linewidth]{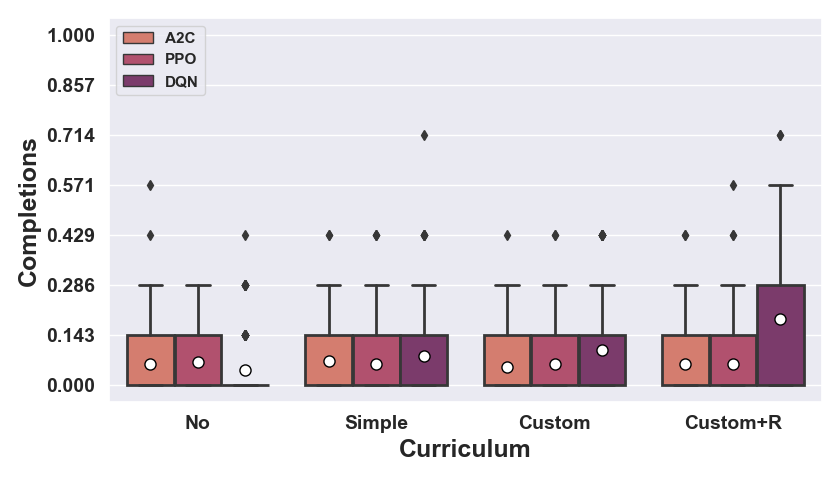}
    \hspace{0.001cm}
    \includegraphics[width=0.49\linewidth]{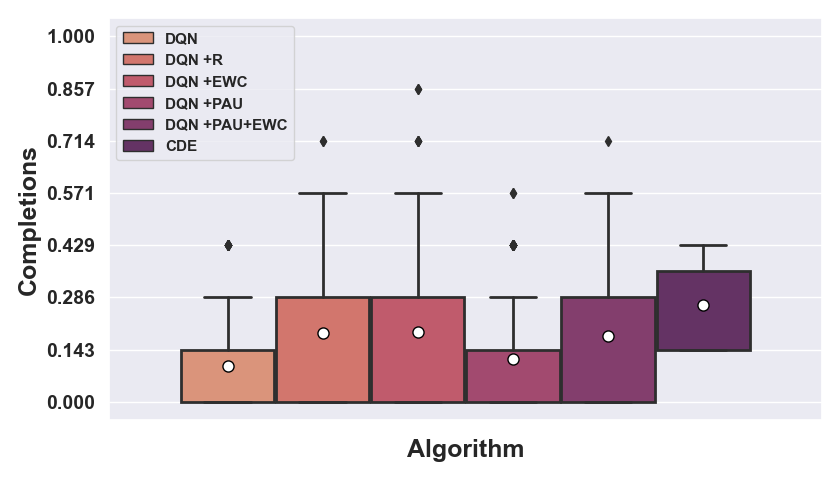}
    \caption{Multiple graphs in a single figure. The first row contains the first two graphs, while the second row contains the last two graphs. Each graph represents different completion rates and scores obtained using various configurations of the DQN algorithm.}
    \label{fig:Performance}
\end{figure*}

 In Figure \ref{fig:Performance}, we present a comparison of various standard reinforcement learning algorithms trained with different curricula on the evaluation environment. The boxplots illustrate that the incorporation of a curriculum significantly enhances the performance of all baselines in terms of scores and completion rates. Specifically, our proposed curriculum, combined with rehearsal, substantially improves the performance of the DQN algorithm, both in scores and completions, compared to a simpler curriculum. This underscores the benefits of a structured learning approach. However, the improvements observed for PPO and A2C are less pronounced, generally remaining within the margin of error. This highlights the necessity of aligning curricula with algorithms that are appropriately suited to the specific task at hand.

The second set of boxplots in Figure \ref{fig:Performance} further extends this analysis by comparing DQN with various enhancements, specifically assessing generalization performance using our custom curricula. Notably, our proposed method, Continual DQN Expansion (CDE), demonstrates superior performance, achieving nearly a 50\% improvement in completion rates compared to the baseline DQN with rehearsal.

In Table \ref{table:curriculum_extended}, we present an extended version where we show why scores are not as informative as the completion rates. 

\begin{table*}[ht]
\caption{Performance comparison of different baselines trained with a range of curricula. The score (lower is better $\downarrow$) and completion rate (higher is better $\uparrow$) are reported. The best score and completion rate are highlighted in bold. We see that the custom curriculum with decomposed tasks (see Figure \ref{fig:main}) achieves the highest performance. Further improvements are observed for the custom curriculum  with rehearsal in the case DQN. }
\centering
\begin{tabular}{lcccccccc}
\toprule
\multirow{2}{*}{\textbf{Method}} & \multicolumn{2}{c}{\textbf{No Curric.}} & \multicolumn{2}{c}{\textbf{Naive Curric.}} & \multicolumn{2}{c}{\textbf{Custom Curric.}} & \multicolumn{2}{c}{\textbf{Custom Curric. (Rehear.)}} \\
\cmidrule(lr){2-3} \cmidrule(lr){4-5} \cmidrule(lr){6-7} \cmidrule(lr){8-9}
 & Score  & Completion  & Score & Completion  & Score & Completion & Score & Completion \\
\midrule
A2C   & 0.34 ± 0.01 & 0.06 ± 0.01 & $0.34 \pm 0.01$  &  $0.07 \pm 0.03$ & $0.35 \pm 0.01$ & $0.05 \pm  0.02$ & $0.34 \pm  0.01$  & $0.06 \pm  0.02$  \\
PPO   & 0.34 ± 0.01 & 0.06 ± 0.02 & $0.35 \pm 0.01$  &  $0.06 \pm 0.02$ & $0.35 \pm 0.01$ & $0.06 \pm  0.02$ & $0.34 \pm  0.01$  & $0.06 \pm  0.03$  \\

DQN   & 0.35 ± 0.01 & 0.04 ± 0.07 & $0.35 \pm 0.01$  &  $0.08 \pm 0.04$ & $0.34 \pm 0.01$ & $0.10 \pm  0.04$ & $ \mathbf{0.31 \pm  0.02}$  & $\mathbf{0.19 \pm  0.09}$  \\

\bottomrule
\end{tabular}

\label{table:curriculum_extended}

\end{table*}

In Table \ref{table:training_timeapp} we display the wall-clock times in seconds for the training of different algorithms and the generation of the environments needed for the Curricula.

\newpage
\begin{table}[ht]
\caption{Training time of each method/curricula in wall-clock time (seconds). Here we only count the effective training time for the algorithm and the cumulative environment generation time for the curricula. The total number of iterations is fixed across methods.}
\centering
\begin{tabular}{lc}
\toprule
\textbf{Method} & \textbf{Training Time (s)} \\
\midrule
A2C & \: \: \: 889 ± \:  64 \\
PPO & \: \: \: 812 ± \: 72 \\
DQN                           & \: \: 1182 ± \: 84   \\
\hline
DQN +EWC \cite{kirkpatrick2017overcoming}   & \: \: 1206 ± \: 80   \\
DQN +MAS \cite{aljundi2018memory}           & \: \: 1198 ± \: 76   \\
\hline

DQN +PAU \cite{delfosse2021adaptive}        & \: \: 2128 ± 102  \\
DQN +CBP \cite{dohare2021continual}         & \: \: 1655 ± \: 96   \\
\hline

DQN +PAU+EWC                               & \: \: 2178 ± \: 89   \\
CDE (ours)                                 & \: \: 3138 ± 138  \\
CDE-1                                 & \: \: 3688 ± 165  \\
CDE-3                                 & \: \: 7433 ± 221  \\
CDE-5                                 & \: 11089 ± 293  \\
CDE-27                                 & \:  13053 ± 342  \\
\toprule
\textbf{Curriculum Generation} & \textbf{Loading Time (s)} \\
\midrule

No Curric.                                 & \: 19864 ± 812  \\
Naive Curric.                                 & \: 12872 ± 599  \\
Custom Curric.                                 & \: 10733 ± 512 \\
Custom Curric. (Rehear.)                                 & \: \:   8826 ± 483 \\
\bottomrule
\end{tabular}

\label{table:training_timeapp}
\end{table}

\begin{figure*}[h]
    \centering
    \includegraphics[width=0.33\linewidth]{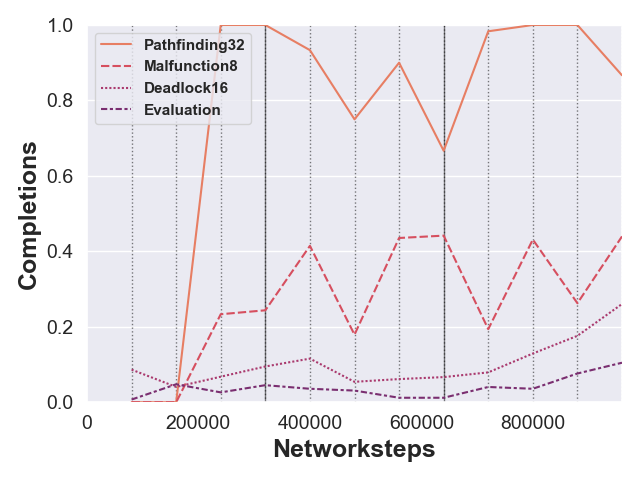}
    \includegraphics[width=0.33\linewidth]{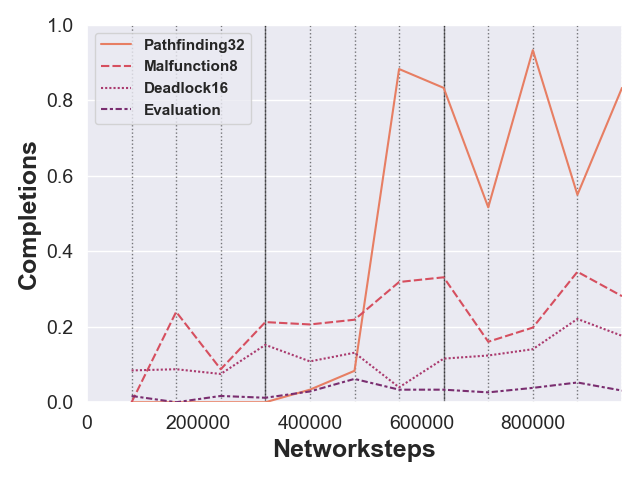}
    \includegraphics[width=0.33\linewidth]{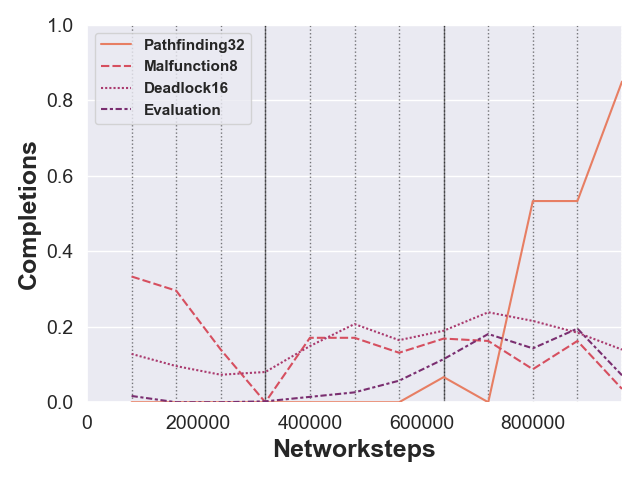}
    \caption{Task ordering impact analysis for DQN baseline across different curricula: PMD, MPD, and MDP (from left to right, top to bottom). Here, P stands for Pathfinding, M for Malfunctions with varying train speeds, and D for Deadlocks. The dotted purple line represents performance on the final test environment, while the orange, red, and pink lines correspond to the Pathfinding, Malfunction, and Deadlock environments, respectively. Vertical black lines indicate transitions to new tasks. The results highlight significant fluctuations in DQN's performance on the test environment depending on the task sequence. A noticeable performance drop is observed when DQN transitions to a new task, suggesting some degree of catastrophic forgetting. Furthermore, DQN appears to overfit the Pathfinding environment, which compromises its generalization capability.}
    \label{fig:Dqn}
\end{figure*}

\textbf{Task Ordering:} Figure \ref{fig:Dqn} explores the impact of task ordering on generalization as well as the extent of catastrophic forgetting across previous tasks for DQN baseline. The results indicate that the performance of DQN on the test environment (the dotted purple line) fluctuates significantly depending on the task sequence. On average, some degree of forgetting is observed when DQN transitions to a new task as evidenced by the drop in performance in the pathfinding, malfunction and deadlocks environment. Nevertheless, DQN seems to overfit in the pathfinding task.  

\begin{figure*}[h]
    \centering
    \includegraphics[width=0.32\linewidth]{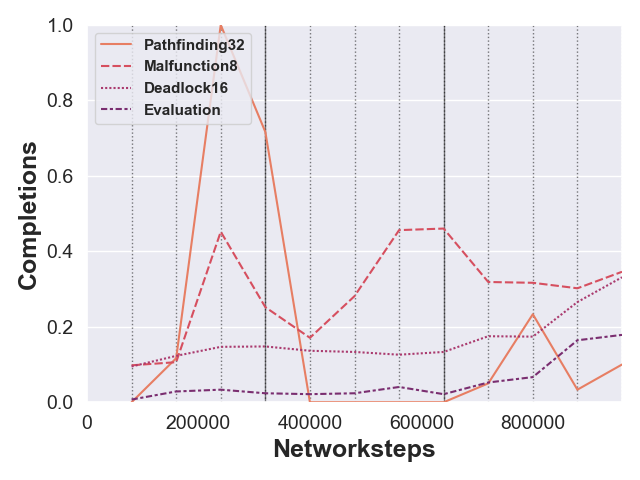}
    \hspace{0.001cm}
    \includegraphics[width=0.32\linewidth]{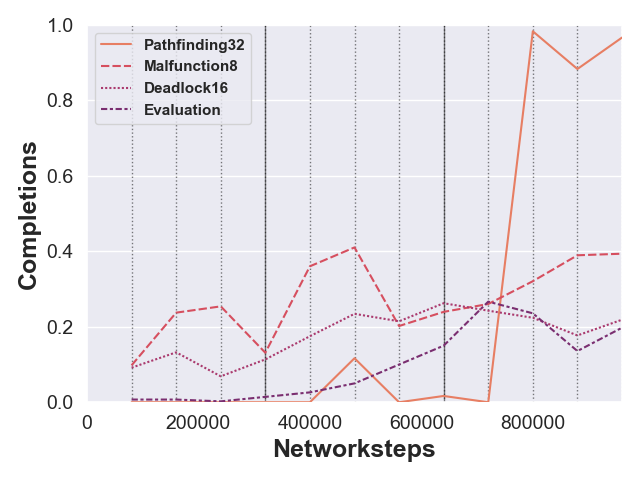}
    \includegraphics[width=0.32\linewidth]{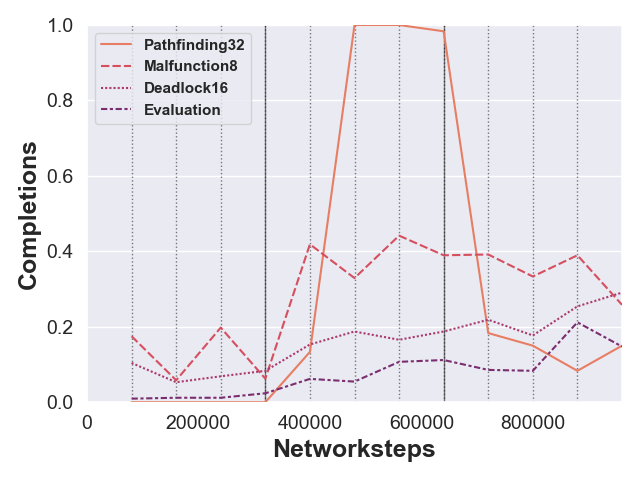}
  
    \caption{Task ordering impact analysis for CDE across different curricula: PMD, MPD, and MDP (from left to right, top to bottom). Here, P stands for Pathfinding, M for Malfunctions with varying train speeds, and D for Deadlocks. The dotted purple line represents performance on the final test environment, while the orange, red, and pink lines correspond to the Pathfinding, Malfunction, and Deadlock environments, respectively. Vertical black lines indicate transitions to new tasks. The results demonstrate that CDE is less sensitive to task ordering, consistently achieving stable performance on the test environment regardless of the task sequence. CDE effectively builds on prior skills to steadily boost zero-shot performance, and resists catastrophic forgetting.}
    \label{fig:CDE}
\end{figure*}

\begin{figure*}[h]
    \centering
    \includegraphics[width=0.24\linewidth]{figures/Subenv_Performance/DQN/DQN_customPMD_compeltions.png}
    \hspace{0.001cm}
    \includegraphics[width=0.24\linewidth]{figures/Subenv_Performance/CDE0.5/CDE_customPMD_completions.png}
    \includegraphics[width=0.24\linewidth]{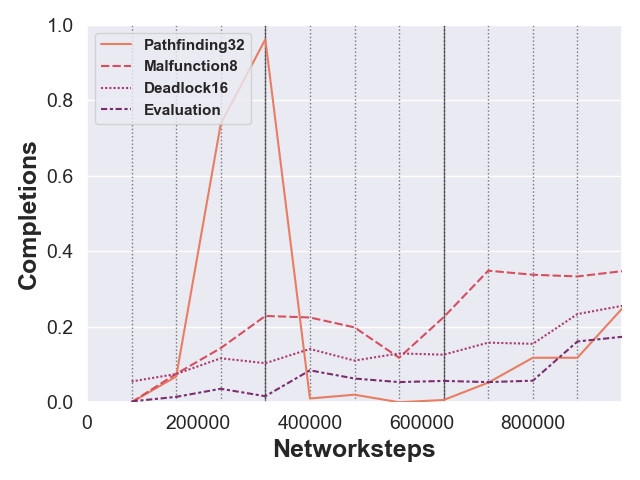}
    \hspace{0.001cm}
    \includegraphics[width=0.24\linewidth]{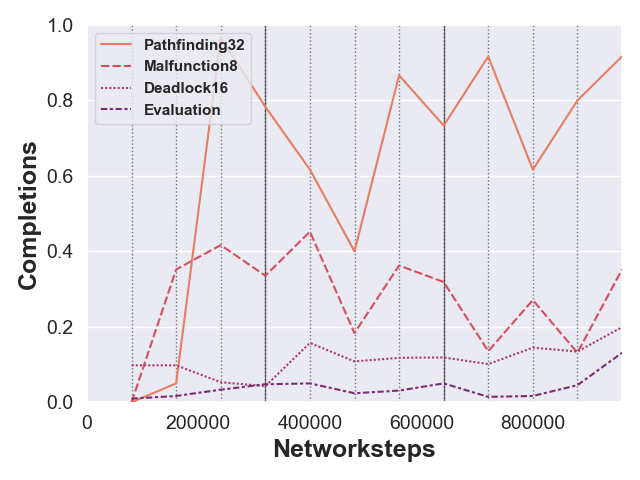}
   \caption{From left to right: skill retention analysis for DQN, CDE, DQN(EWC), and DQN(PAU) across different environments. The orange line represents performance on the Pathfinding environment, the dotted red line indicates the Malfunction environment, the dotted pink line corresponds to the Deadlock environment, and the dotted purple line represents generalization performance on the evaluation environment. Vertical black lines mark transitions to new tasks, which follow the order: Pathfinding, Malfunctions with varying train speeds, and Deadlocks. While DQN maintains the highest performance on the Pathfinding task, suggesting potential overfitting and reduced plasticity, CDE shows a balanced approach, effectively retaining core skills while adapting to new challenges, as indicated by steady improvements in generalization performance (purple line). Similar trends are observed for DQN(EWC) and DQN(PAU), but they achieve lower generalization performance compared to CDE. See the main body for detailed results with error margins.}
    \label{fig:multiple_graphs}
\end{figure*}

In contrast, Figure \ref{fig:CDE} demonstrates that CDE is considerably less sensitive to task ordering, consistently achieving similar performance on the final test environment, regardless of the task sequence used. For instance, when comparing the PMD (Pathfinding, Malfunctions, and Deadlocks) and MPD (Malfunctions, Pathfinding, and Deadlocks) curricula, CDE effectively builds on previously acquired skills, gradually improving completion rates in zero-shot test environments. 
\\\\\\\\
\textbf{Skill Retention:} 
Figure \ref{fig:multiple_graphs} illustrates the retention of skills across previous tasks for DQN, DQN(EWC), DQN(PAU), and CDE. The orange line represents performance on the Pathfinding environment, the dotted red line corresponds to the Malfunction environment, the dotted pink line indicates the Deadlock environment, and the dotted purple line denotes the evaluation environment. 

For the Pathfinding task (orange line), a noticeable performance drop is observed for both CDE and DQN(EWC) when transitioning to subsequent tasks, as marked by the vertical black line. We hypothesize that this suggests that while these methods adapt to new skills, some trade-off in retaining earlier skills occurs. In contrast, DQN maintains the highest performance on the Pathfinding environment, indicating a tendency to overfit this task. This overfitting hampers the model's plasticity, making the acquisition of new skills more challenging and resulting in poorer performance on the generalization environment (dotted purple line).

Although CDE experiences a performance drop in the Pathfinding task when transitioning to new tasks, this decline can be likely attributed to the algorithm’s ability to adapt and acquire new skills, such as handling malfunctions and deadlocks. CDE effectively retains core skills while simultaneously learning to manage newer challenges, such as varying train speeds. Consequently, the single-agent Pathfinding skill becomes less critical in subsequent tasks, as evidenced by the steadily increasing performance in the generalization environment (purple dotted line).This highlights CDE's flexibility in balancing stability and plasticity, retaining key skills from past tasks while discarding less relevant ones. Compared to other methods, it achieves a more effective stability–adaptability trade-off.

\newpage
\section*{Limitations and Future Works}

While our approach demonstrates promising results, there are several points to consider:

1. \textbf{Computational Overhead:} Introducing more subspaces in the CDE framework inevitably leads to increased computational overhead. Although we partially mitigate this issue by limiting the number of subspaces and rotating the acting network during training, the computational demands remain significant.

2. \textbf{Balancing Catastrophic Forgetting and Adaptability:} In this work, our goal extended beyond merely addressing catastrophic forgetting; we aimed to develop a system that retains relevant "skills" from previous tasks and adapts to new tasks when necessary. This approach differs from the traditional continual learning paradigm, where the primary objective is to preserve as much knowledge as possible from previous tasks while learning new ones. A core challenge for continual learning research in understanding and better categorizing different continual learning  scenarios and thus creating more holistic systems capable of tackling a variety of open-world challenges.

3. \textbf{Longer Curricula and Task Non-Stationarity:} This work primarily focuses on a single type of non-stationary continual learning scenario. It remains an open question how the system would perform with much longer curricula or in scenarios where non-stationarity is expressed through changing tasks like in ATARI games. Future investigations should consider these factors to enhance the system's adaptability across a wider range of continual learning settings.

4. \textbf{Comparison to Operations Research Approaches:} Despite achieving good results on the train scheduling problem, our system still exhibits a performance gap compared to traditional operations research methods. There are still open questions on how to best solve this problem while ensuring that the system remains highly adaptable and capable of generalizing across diverse tasks.

\end{document}

%% file: math_commands.tex
\usepackage{amsmath,amsfonts,bm}

\def\eqref#1{equation~\ref{#1}}

\def\1{\bm{1}}

\DeclareMathAlphabet{\mathsfit}{\encodingdefault}{\sfdefault}{m}{sl}
\SetMathAlphabet{\mathsfit}{bold}{\encodingdefault}{\sfdefault}{bx}{n}

